\documentclass[conference]{IEEEtran}
\IEEEoverridecommandlockouts
\usepackage{cite}
\usepackage{amsmath,amssymb,amsfonts}
\usepackage{algorithmic}
\usepackage{graphicx}
\usepackage{textcomp}
\usepackage{xcolor}

\usepackage{multirow}
\usepackage[hyphens]{url}
\usepackage{verbatim}
\usepackage{tabularx}
\usepackage{booktabs}
\usepackage{comment}

\usepackage{tikz}

\newcommand\copyrighttext{%
  \footnotesize \textcopyright \the\year{} IEEE. Personal use of this material is permitted. Permission from IEEE must be obtained for all other uses, including reprinting/republishing this material for advertising or promotional purposes, collecting new collected works for resale or redistribution to servers or lists, or reuse of any copyrighted component of this work in other works.}

\newcommand\copyrightnotice{%
\begin{tikzpicture}[remember picture,overlay]
\node[anchor=south,yshift=10pt] at (current page.south) {\fbox{\parbox{\dimexpr0.75\textwidth-\fboxsep-\fboxrule\relax}{\copyrighttext}}};
\end{tikzpicture}%
}

\usepackage{color}
\newcommand{\new}[1]{#1}

\def\BibTeX{{\rm B\kern-.05em{\sc i\kern-.025em b}\kern-.08em
    T\kern-.1667em\lower.7ex\hbox{E}\kern-.125emX}}
\begin{document}

\title{A Multimodal Architecture for Endpoint Position Prediction in Team-based Multiplayer Games}


\author{\IEEEauthorblockN{Jonas Pech\'e}
    \IEEEauthorblockA{\textit{Computer Graphics} \\
    \textit{Johannes Kepler University Linz}\\
    Linz, Austria \\
    j\_peche@wargaming.net}
\and
    \IEEEauthorblockN{Aliaksei Tsishurou}
    \IEEEauthorblockA{\textit{Independent Researcher} \\
    Wroclaw, Poland \\
    aliakseitsishurou@gmail.com}
\and
    \IEEEauthorblockN{Alexander Zap}
    \IEEEauthorblockA{\textit{DS Research} \\
    \textit{Wargaming}\\
    Berlin, Germany \\
    a\_zap@wargaming.net}
\and
    \IEEEauthorblockN{G\"unter Wallner}
    \IEEEauthorblockA{\textit{Computer Graphics} \\
    \textit{Johannes Kepler University Linz}\\
    Linz, Austria \\
    guenter.wallner@jku.at}
}

\maketitle
\copyrightnotice

\begin{abstract}
\new{Understanding and predicting player movement in multiplayer games is crucial for achieving use cases such as player-mimicking bot navigation, preemptive bot control, strategy recommendation, and real-time player behavior analytics. However, the complex environments allow for a high degree of navigational freedom, and the interactions and team-play between players require models that make effective use of the available heterogeneous input data.} This paper presents a multimodal architecture for predicting future player locations on a dynamic time horizon, using a \emph{U-Net}-based approach for calculating endpoint location probability heatmaps, conditioned using a multimodal feature encoder. The application of a multi-head attention mechanism for different groups of features allows for communication between agents. In doing so, the architecture makes efficient use of the multimodal game state including image inputs, numerical and categorical features, as well as dynamic game data. Consequently, the presented technique lays the foundation for various downstream tasks that rely on future player positions such as the creation of player-predictive bot behavior or player anomaly detection.
\end{abstract}


\section{Introduction}
    \label{sec:introduction}

Predicting the future position of players in team-based video game environments is important for a variety of tasks such as AI-based decision-making, strategy optimization of bots, and real-time player behavior analysis. However, such predictions can be challenging, especially in dynamic game environments with complex interactions between multiple entities. 

Besides applications in video games, the task of location prediction has found wide-ranging interest in domains such as robotics~\cite{9920012}, autonomous driving~\cite{9349962}, and sports analytics~\cite{hauri2020multimodal}. However, unlike in many real-life scenarios where navigation is constrained by conventions and rules, video games such as \emph{World of Tanks} (\emph{WoT})~\cite{wot} allow \new{for more potential movement options} when it comes to navigating their virtual terrain. Thus, to solve the task of location prediction in an environment in which multiple players constantly react to each other, and where a variety of different game information is available at every time point, an approach with a good representation of multiple agents and multimodal information is required. 

As such we decided on an image-to-image prediction approach, to be able to make use of commonly available top-down map data in video games. The prediction format of our technique is a heatmap of probabilities as visualized in Figure~\ref{fig:prediction1}. This can be a more suitable representation for many tasks that require only the endpoints (and not the trajectory leading towards them). \new{Such tasks include, for example, the strategic placement of bots, improvement of long-range aiming by opponent anticipation, and selection of points-of-interest during a match}.
Moreover, a heatmap showing probabilities also offers a more complete representation as opposed to a fixed number of sampled endpoints.

While there are many such image-to-image prediction approaches available~\cite{mangalam2020goalswaypointspaths,ren2021safetyawaremotionpredictionunseen, izquierdo2022vehicletrajectorypredictionhighways}, there is an absence of using them in the video games domain. In this paper we describe how we use a \emph{U-Net}~\cite{ronneberger2015unet} as a baseline for this task and apply various improvements and strategies to further enhance its performance. These techniques were gathered and adapted from various fields. To model the spatial and temporal dependencies and predict endpoint probabilities accurately, we propose a multimodal encoder that can leverage heterogeneous inputs such as images, per-vehicle data, global categorical and numerical data, as well as streams of dynamic data (e.g., positions). To compensate for the lack of intermediate trajectory prediction, we condition the output with a variable prediction horizon.

We then evaluate how the encoder components affect the prediction accuracy through an ablation study and assess how various strategies can improve performance. For this, a vast dataset with diverse environments across different levels of urbanization is used.

\section{Related Work}
    \label{sec:relatedwork}

Many different approaches for predicting endpoints or trajectories with different strengths and use cases have been proposed to date. A straightforward implementation could predict two or three-dimensional endpoint positions as a regression task, but this approach has proven to be ineffective and limited. For example, Bishop~\cite{370fbeadb5584ba9ab2938431fc4f140} already discussed the limitations of predicting continuous values using regression back in 1994. Alternatively, a trajectory can also be modeled as a sequence-to-sequence task as done, for instance, by Ivanovic and Pavone~\cite{ivanovic2019trajectronprobabilisticmultiagenttrajectory} for vehicle trajectories in traffic situations, or by Sandro et. al.~\cite{hauri2020multimodal} who used a recurrent network for predicting the movement of NBA players. Through the use of variational autoencoders, multimodality can be achieved by sampling multiple trajectories such as done by \emph{PECNet}~\cite{mangalam2020journeydestinationendpointconditioned}. However, in use cases with a high degree of freedom such as in multiplayer video games, heatmap-based predictions might be more suitable. Heatmaps as prediction modalities are used in a wide variety of areas, including human movement~\cite{mangalam2020goalswaypointspaths} and human pose estimation~\cite{chen20222dhumanposeestimation,zheng2023deeplearningbasedhumanpose}, with often better results than other approaches. For instance, Mangalam et. al.~\cite{mangalam2020goalswaypointspaths} presented a \emph{U-Net}-based approach for way- and end-point prediction and extended it with subsequential trajectory prediction. However, they did not further explore additional context conditioning or multi-agent environments.

Especially in strategic scenarios, where decisions made by agents depend on the other agents, some form of communication is crucial. Social pooling as introduced by Alahi et al.~\cite{sociallstm} and its various improvements put a strong focus on agent behavior and interactions by merging the hidden states of a recurrent component based on their spatial distance, but are less suited for large scale or long-term predictions. Other frequently used approaches to allow sharing information between agents include attention mechanism and graph neural networks such as the spatial and temporal modeling approach proposed by Huang et al.~\cite{9010834}. Yeh et al.~\cite{8953867}, on the other hand, combined a graph neural network with variational recurrent layers to predict movement in sports games, notable also exploring generalization by using counterfactual situations as input.

\section{Method}
    \label{sec:method}

The goal of our method is to estimate a heatmap of the most likely positions a specific target vehicle will occupy at a given time horizon. To be able to efficiently combine various inputs, our proposed approach leverages a \emph{U-Net}~\cite{ronneberger2015unet} and conditions it with global and vehicle state information. In this way, the visual top-down game map, enriched with additional vehicle information, can be efficiently used as an input to the encoder of the \emph{U-Net}, \new{making it more suited than decoder-only architectures}. We then introduce a generic multimodal encoder to process both global state and vehicles. Two stacked attention layers then establish the required communication between the agents, allowing the model to learn agent interactions and dynamics. The architecture thus consists of the following three main elements: (1) an image encoder and decoder which together form the \emph{U-Net} backbone, (2) a multimodal feature encoder combining recurrent, embedding, and numerical data, and (3) a conditioner, embedding the combined features into the image decoder.

\subsection{Data}
    \label{sec:data}

To better understand the format and content of the multimodal data, the data preparation is presented first. For training and evaluation we used a dataset consisting of 2,190,000 battles from \emph{WoT}, each with 30 vehicles and an average length of 5 minutes. Each battle is a stream of game states, sampled at intervals of 15 seconds. They are roughly equally split into 29 maps of varying characteristics, environments, and a size of 1,000 $\times$ 1,000 m in side length. The data was collected over a time period of around 810 days and weighted towards higher tank tiers and thus battles from higher experienced players. \new{While we focus on \emph{WoT} battles in this paper, we expect this approach to be applicable to similar team-based games or use cases which have a strong emphasis on positional strategical decision-making. Also, this approach is applicable to game which work with continuous gameplay rather than discrete battles.}

Each game state describes a multimodal situation and consists of the following information: target position, image input, global context (both static and dynamic), per-vehicle context, and per-vehicle history. This information is described in more detail in the following sections.

\subsubsection{Target Position Representation}
    \label{sec:target_generation}

To train the \emph{U-Net}, the target vehicle position needs to be transformed into a 2D image representation first. Rather than using a one-hot encoding of the target position (which would result in very sparse features), a Gaussian ellipsoid is used to encode both position and velocity of the vehicle:
\begin{align}
    K(x, y) &= \frac{G_y(y) \cdot G_x(x)}{\max(G_y(y) \cdot G_x(x))}
    \label{eq:gaussian}
\end{align}

with 
\begin{align}
    G_x(x) &= e^{-\frac{x^2}{2\sigma_x^2}} \\ 
    G_y(y) &= e^{-\frac{y^2}{2\sigma_y^2}} \cdot M(y) \\ 
    M(y) &=
    \begin{cases} 
        \left( \frac{y}{y_{\text{mod}}} \right)^3, & |y| < y_{\text{mod}} \\
        1, & \text{otherwise}
    \end{cases}
\end{align}

\begin{figure}[t]
    \centering
    \begin{minipage}[t]{0.318\textwidth}
        \centering
        \includegraphics[width=\linewidth]{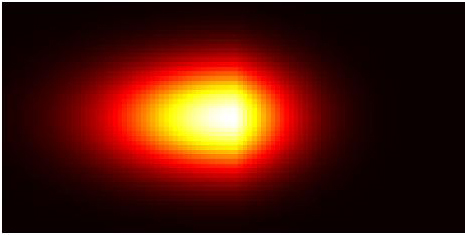}
        \caption{The stretched position-velocity kernel, at 50\% maximum speed, a kernel size of 50 and a sigma of 8, rotated towards the right. Kernel intensity: 0~\includegraphics[height=0.6\baselineskip]{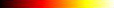}~1}
        \label{fig:kernel}
    \end{minipage}
    \hfill
    \begin{minipage}[t]{0.163\textwidth}
        \centering
        \includegraphics[width=\linewidth]{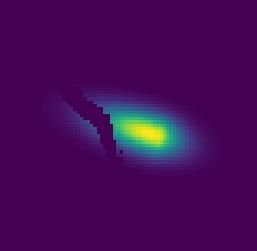}
        \caption{An example target for a moving vehicle with inaccessible or obstructed areas masked out.}
        \label{fig:gaussian_target}
    \end{minipage}
\end{figure}

\noindent The vertical Gaussian's size and sigma ($G_y(y)$) is stretched with up to twice the current normalized vehicle speed: 
\begin{equation}
1 + k \cdot \min\left(1, \frac{\text{current\_speed}}{\text{max\_speed} + \epsilon}\right)
\end{equation}

\noindent attenuated by a cubic multiplier at the upper fraction to simulate the technical limitation (maximum speed) of the vehicle. A stretching factor $k$ of 2.0 is chosen to get a distribution in the range $(-\text{max\_speed}, \text{max\_speed})$. Then the kernel $K(x, y)$ is normalized to 1.0 to get a probability distribution. Figure~\ref{fig:kernel} visualizes this kernel.
By rotating the ellipsoid in the direction the vehicle is moving, the model is guided with additional information. This approach was inspired by Gilles et al.~\cite{gilles2022thomastrajectoryheatmapoutput}, who used Gaussian distributions centered on the target position. Zheng et al.~\cite{zheng2023deeplearningbasedhumanpose} also used Gaussian kernels as targets for heatmap-based pose estimation approaches.

While the velocity is not directly predicted or included in the loss or evaluation metrics in Section~\ref{sec:evaluation}, it helps the model fit a better uncertainty distribution.

In addition, inaccessible or obstructed areas are masked out, guiding the model to predict correct probability distributions rather than just ignoring those pixels as part of a loss mask. Figure~\ref{fig:gaussian_target} visualizes this target for a moving vehicle, partially obscured by an obstacle.

\subsubsection{Image Input}
    \label{sec:image_input} 

\begin{figure}[t]
    \centering
    \includegraphics[width=0.92\linewidth,trim={0cm 4cm 0cm 0cm},clip]{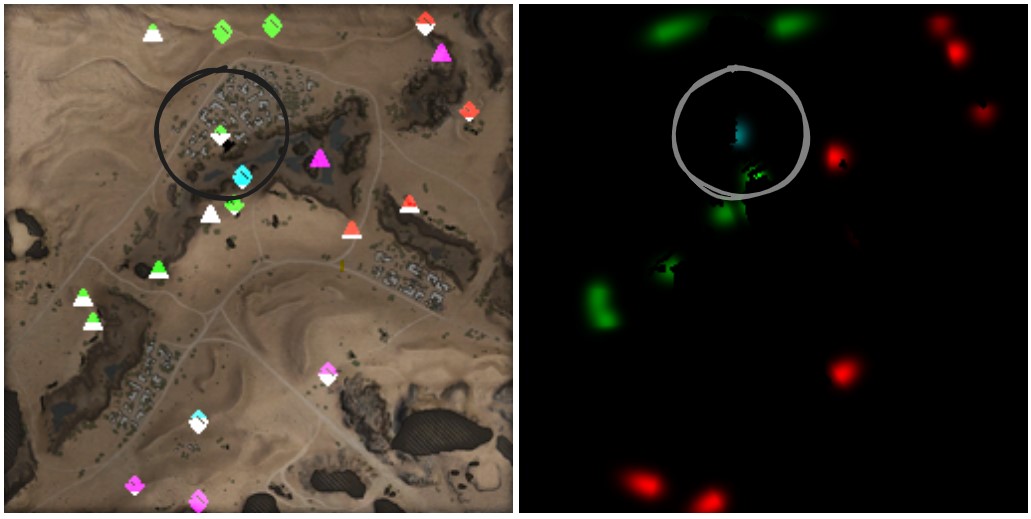}
    \caption{The multidimensional input image, separated into RGB (left) and the Gaussian ellipsoid data (right) for the target, ally team, and enemy team. The target vehicle is highlighted in this visualization with a black and white circle respectively.}
    \label{fig:image_input}
\end{figure}

We use the RGB environment color, similar to what the player would see on their minimap, as the image input. This static data encodes information about the terrain type, obstacles, accessible areas, coverage, and foliage. To enhance this representation, we extend the three-channel RGB image with three additional feature maps that capture vehicle position and velocity information in a float-encoded image (same size as the RGB image): one for the current vehicle whose endpoint we aim to predict, one for allied vehicles, and a third for enemy vehicles, as visualized in Figure~\ref{fig:image_input} (right). For the team-based information maps, the features of all vehicles at a position are summed, allowing for values greater than 1.0 for vehicles close to each other. For the encoding we once again use Gaussian ellipsoids as in Section~\ref{sec:target_generation}.

Furthermore, the information density can be further increased by adding additional information (vehicle type, health, visibility to opponents) to the RGB image. For this purpose, two possible solutions were investigated: (A-1) several Gaussian ellipsoids are stacked into additional feature channels, using the intensity to encode a continuous or boolean value. (A-2) the additional features are encoded as icons and added to the existing RGB data of the mini-map. 
Visibility is encoded as 100\% opacity if the vehicle is visible or 50\% opacity if the vehicle is invisible to the opponent. Health is encoded by vertically inverting the color of the icon proportional to the missing health fraction. We compare both approaches in Section~\ref{sec:evaluation} and set it in relation with the baseline that lacks those features completely. Figure~\ref{fig:image_input} (left) shows a cross-section of this multidimensional image, with the additional features rendered on top of the RGB data (as described for A-2).

\subsubsection{Global Context}

Global context information includes the current time since the beginning of the round, the current map, the game mode, as well as a team-wise aggregate over the vehicle health and damage applied to enemy vehicles. To get a fixed-size vector independent of the vehicle count, the health and damage dealt are averaged over the five primary vehicle types (heavy, medium, light, tank destroyers, and artillery) for each of the two teams. If vehicle types are not used, zero values are used. The global context contains the prediction horizon as well, which we refer to as conditioning.

\subsubsection{Vehicle Context}

The vehicle context consists of the categorical features team, vehicle type, role, and name. It also includes numerical features such as position, speed, current health in percent, the damage dealt until that point, and the time since it was last seen by the enemy team. 
In addition, the total battle count and current rating of player in ranked matchmaking is included as the behavior depends on the skill and experience level. We round up both, rating and number of battles, to the nearest hundred to both support anonymity and prevent the model from remembering specific player patterns.

\subsubsection{Vehicle History}

To capture data dynamics, the position, spatial orientation of the hull and turret of the tank, as well as health over time are represented as a multidimensional time series. This allows the model to understand and predict based on vehicle behavior beyond the current static state. For our dataset, history is provided as up to 15 samples with a 15-second step size.

\subsubsection{Data Sampling}

The data is provided in 15-second steps, and samples are drawn uniformly randomly from the set of available steps, excluding the last seconds spanned by the prediction horizon. The horizon is a randomly sampled value between 1 and 6 steps (i.e. 15 and 90 seconds). Because it is very common for a vehicle to not move at all, we resampled the dataset to only train from vehicles which moved at least 6\% of their maximum speed (defined as the velocity corresponding to the maximum slope in the speed distribution). To prevent only learning moving behavior, we additionally sample a random vehicle in 10\% of cases. That way, more focus is put on challenging scenarios, while still including situations where the vehicle is expected to remain in the same location. Both values were chosen based on internal requirements and should be adjusted to the expected inference situation distribution.


\subsection{Architecture}

\begin{figure}
    \centering
    \includegraphics[width=1.0\linewidth,trim={3cm 3cm 3cm 3cm},clip]{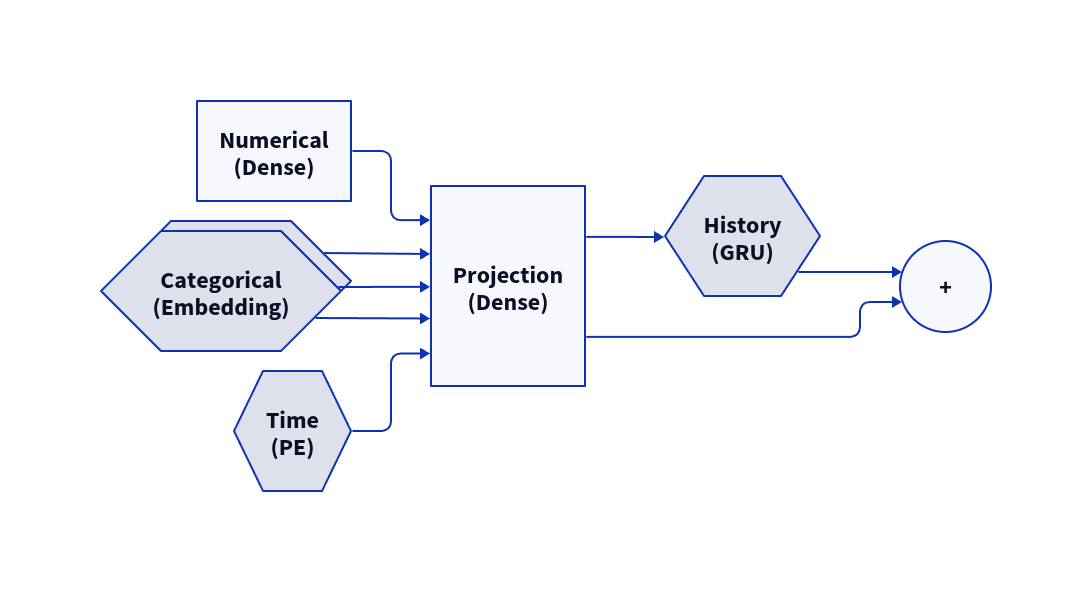}
    \caption{Our numerical-categorical encoder, the primary encoding component for vehicles and global data.}
    \label{fig:encoder}
\end{figure}

In the following we describe how the data presented in Section~\ref{sec:data} is encoded. The primary building block is the numerical-categorical encoder (NCE) as visualized in Figure~\ref{fig:encoder} and which encodes the multimodal input into one embedding vector. The encoder is defined as:
    \begin{align}
    \label{eq:encoder}
        h &= \phi \left( \text{BatchNorm1d} \left( \left[ h_{\text{num}}, \, \bigoplus_{i=1}^{n} h_{\text{cat}_i}, \, h_{\text{time}} \right] W + b \right) \right) \nonumber \\
        z &= h + \text{GRU}(x_\text{dyn}, h)
    \end{align}

\noindent It combines a dense layer and batch normalization~\cite{ioffe2015batchnormalizationacceleratingdeep} for numerical data $h_\text{num}$, several concatenated categorical embeddings $h_\text{cat}$, and a time embedding $h_\text{time}$. One final dense layer with a second batch normalization maps the concatenated parts into a single embedding $z$. A recurrent network over a skip connection then uses $z$ as the initial state, encoding the dynamic features $x_\text{dyn}$.

One such numerical-categorical encoder is used for the global data (for which the recurrent network is not used), one for the target vehicle, and a shared one for all remaining vehicles, i.e.: $z_\text{global} = E_\text{global}(x_\text{global})$, $z_\text{target} = E_\text{target}(x_\text{target})$, $z_\text{i} = E_\text{vehicle}(x_\text{i})$.

The target vehicle uses a separate encoder to encourage differentiating between context and target vehicles with slightly different patterns to learn. This separation also allows using two different sized encoders, e.g., to better match inference time budget constraints in real-time applications. The time input here is the time since the vehicle was last visible. The global encoder $E_\text{global}$ does not make use of the recurrent component due to the lack of global dynamic data series. One final dense layer maps the concatenated final vehicle embedding from Section~\ref{sec:attention}, the global embedding, and the prediction horizon encoding using the time encoder from Section~\ref{sec:time} to a single final embedding vector:
\begin{equation}
z_{\text{final}} = \phi\left( \left[ h_{\text{cross}}, h_{\text{global}}, h_{\text{horizon}} \right] W + b \right)
\end{equation}

\begin{figure}
    \centering
    \includegraphics[width=1.0\linewidth,trim={0cm 0cm 0cm 0cm},clip]{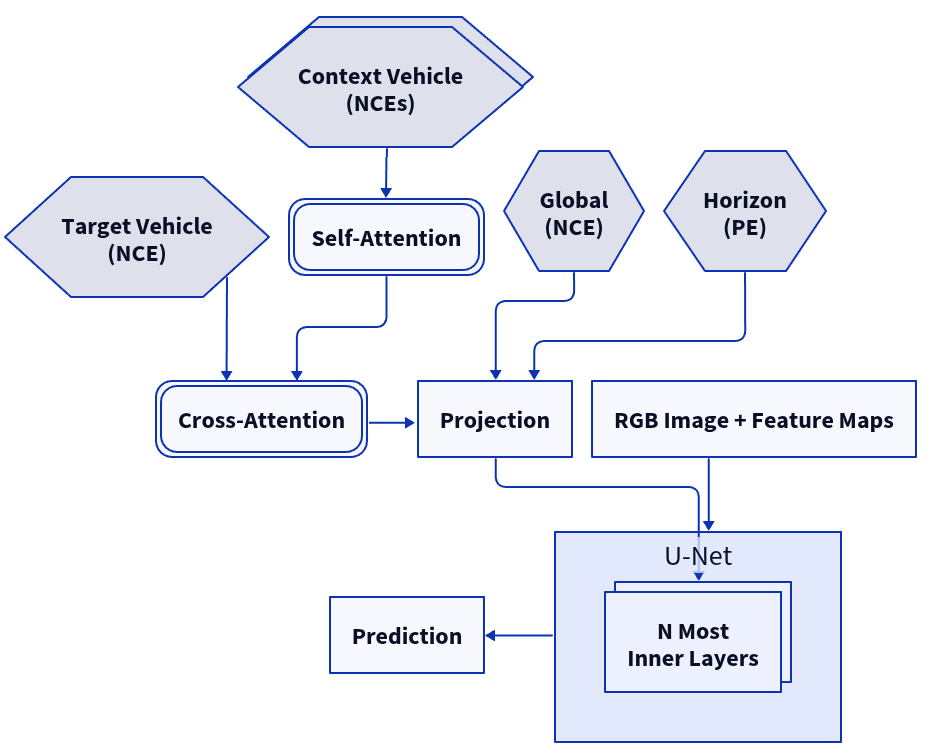}
    \caption{The higher level architecture, visualizing the global, target, and context vehicle numerical-categorical encoders as well as horizon and attention components. The final embedding then conditions the hidden layers of the \emph{U-Net}.}
    \label{fig:architecture}
\end{figure}

\noindent This vector is then used to condition the \emph{U-Net}. The final architecture is outlined in Figure~\ref{fig:architecture}. 

\subsubsection{Time and Horizon Encoding}
    \label{sec:time}

For encoding the current time and prediction horizon, a positional embedding is used to preserve the relation between temporally close samples. Specifically, we used the Sinusoidal Positional Encoding (PE) presented by Rombach et al.~\cite{rombach2022highresolutionimagesynthesislatent} and defined as:
\begin{equation}
    \label{eq:pe}
    PE_{\text{pos},2i} = \sin\left(\frac{\text{pos}}{M^{2i/d}}\right),
    PE_{\text{pos},2i+1} = \cos\left(\frac{\text{pos}}{M^{2i/d}}\right)
\end{equation}

\noindent PE was initially suggested in the context of transformers by Vaswani et al.~\cite{vaswani2023attentionneed} as it has been shown to perform equally to a learnable embedding. We picked the maximum possible time and horizon for $M$ (10,000 and 6 respectively), and a dimension of $8$.

\subsubsection{Dynamic Feature Encoding}

Using a recurrent layer, all dynamic features such as position, health, etc. are aggregated into one vector. We decided on a uni-directional GRU~\cite{cho2014learning} cell with bias, but initialized the hidden state with the encoded vehicle state $z_\text{i}$ or $z_\text{target}$ from above, projected to the hidden dimension with a dense layer. This allows the recurrent encoder to consider vehicle characteristics during encoding~\cite{CondRNN}. The final output is then merged back using a skip connection.

\subsubsection{Vehicle Attention}
    \label{sec:attention}

Using the shared encoder $E_\text{vehicle}$, each vehicle is encoded as a vehicle embedding
\begin{equation}
Z = [ z_0, z_1 ... z_n]
\end{equation}

\noindent which forms the input matrix of a multi-head self-attention mechanism
\begin{equation}
Z_\text{self-attn} = Attention(Z, Z, Z)
\end{equation}

\noindent that allows the model to learn relationships and patterns between vehicles. We use the same architecture as proposed by Vaswani et al.~\cite{vaswani2023attentionneed}, including bias but excluding the normalization layer. This layer was moved before the projection layer as suggested by Xiong et al.~\cite{xiong2020layernormalizationtransformerarchitecture}. The output from this attention module is an embedding per vehicle, enriched with information about other vehicles.

Then cross-attention is used, with the query vector being the encoded target vehicle $z_\text{target}$, and the key and value being the other, self-attended vehicles $Z_\text{self-attn}$:
\begin{equation}
z_\text{cross} = Attention(z_\text{target}, Z_\text{self-attn}, Z_\text{self-attn})
\end{equation}

\begin{figure}
    \centering
    \includegraphics[width=0.6\linewidth]{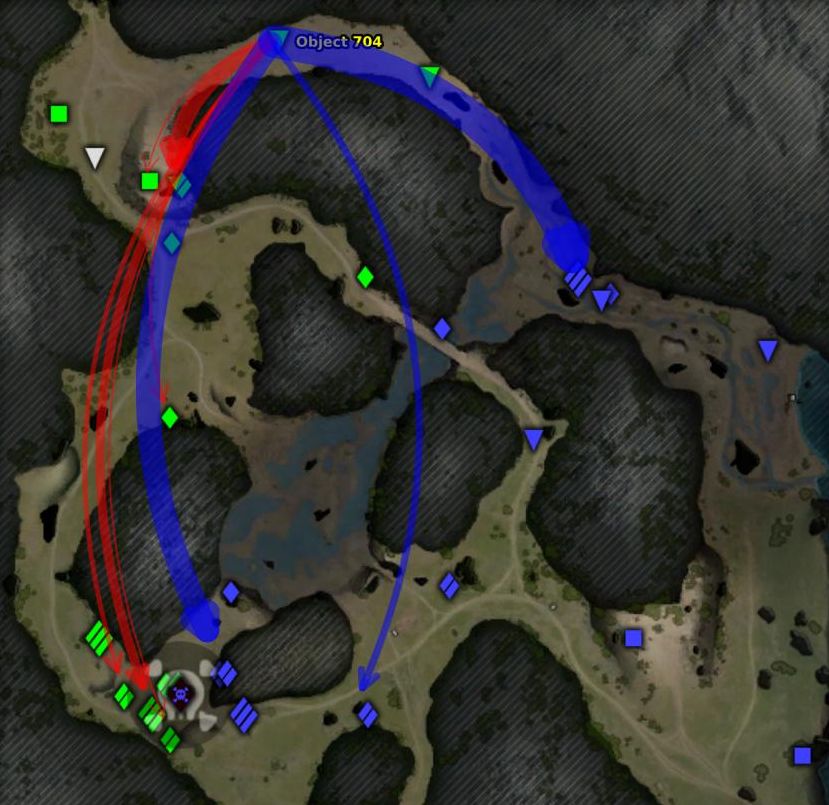}
    \caption{The attention weights of the cross-attention layer visualized as arrows.}
    \label{fig:attention_weights}
\end{figure}

\noindent Hence, the result is a single embedding representing the influence of other vehicles on the target vehicle. A positive side effect of the attention mechanism is interpretability through its weight matrix, which is trained to represent feature importance. While not directly proportional to actual vehicle importance, it still provides interesting details on vehicle interactions. Figure~\ref{fig:attention_weights} shows the cross-attention weights as arrows between target vehicles and all others. It can be seen that the model considers some vehicles as more important for the decision-making of the target vehicle than others.

Interpretability can also be relevant for downstream tasks such as anomaly detection, highlight detection, bot AI, player assistants and coaches, and similar.

\subsubsection{Image Encoder-Decoder}

For both the image encoding and the endpoint position prediction decoder, an image encoder-decoder is used. For the backbone we use \emph{U-Net++}~\cite{zhou2018unetnestedunetarchitecture}, an improvement over the classical \emph{U-Net}~\cite{ronneberger2015unet} architecture by adding nested and dense skip connections. For the encoder we picked \emph{EfficientNet}~\cite{tan2020efficientnetrethinkingmodelscaling} (B2 as a tradeoff between size and capacity) for both simplicity and the ability to use pre-trained models (trained on the \emph{ImageNet}~\cite{ILSVRC15} dataset). \new{We use a convolutional architecture over transformer-based models to reduce computational costs during inference.}
We used the well-maintained segmentation models repository by Iakubovskii~\cite{Iakubovskii:2019}, providing an extensive list of pre-trained models for easy testing and reproducible results. Using a pre-trained backbone reduces required training time even though the task differs a lot from image classification used for backbone pretraining.

We then condition the final embedding vector $z_\text{final}$ obtained by the encoders to the last $n$-hidden layers, with $n$ being a hyperparameter to be evaluated (for our chosen image encoder and use case, a value of 1 is used as we did not notice significant differences with higher values):
\begin{equation}
h_i' = \text{Attention}\left( h_i, z_{\text{final}} \otimes \mathbf{1}, z_{\text{final}} \otimes \mathbf{1} \right), i = 1, 2, \dots, n
\end{equation}

\noindent This approach was inspired by Rombach et al.~\cite{rombach2022highresolutionimagesynthesislatent}, who applied a cross-attention mechanism  to the flattened hidden states of the \emph{U-Net}. To better respect the pre-trained encoder part of the \emph{U-Net}, we only apply this conditioning to the decoder side.

To adapt pre-trained model weights from 3D to higher-dimensional input images, we make use of a weight inflation strategy. This expands the original 3D weight tensor \( W \in \mathbb{R}^{C_{\text{out}} \times C_{\text{in}} \times D \times H \times W} \) into a higher-dimensional tensor \( W' \in \mathbb{R}^{C_{\text{out}} \times C_{\text{in}} \times D' \times H' \times W'} \). The expansion is achieved by repeating the original weights along the new dimensions as needed. This method ensures that the pre-trained weights are reused in the new higher-dimensional context.

\subsection{Prediction}

As finding an appropriate loss function is not trivial for 2D probability distributions, several common loss functions proposed in different fields have been evaluated, \new{each resulting in a slightly different definition of pixel-wise probability}:

\begin{itemize}
    \item \textbf{Binary Cross-Entropy (BCE)}: The most straightforward loss is the BCE between target and prediction probabilities, defined as $\mathcal{L}_{\text{BCE}} = - \sum_{u, v} P(u, v) \log \hat{P}(u, v) + (1 - P(u, v)) \log(1 - \hat{P}(u, v))$.
    \item \textbf{Mean Squared Error (MSE)}: Measures the average squared difference between predicted and actual values: $\mathcal{L}_{\text{MSE}} = \frac{1}{N} \sum_{i=1}^{N} (y_i - \hat{y}_i)^2$. While commonly used in regression tasks, it can also be applied to probabilistic outputs, albeit it is expected to struggle with sharp probability distributions.
    \item \textbf{Focal Loss}: The focal loss puts focus on the target region and generally works better with unbalanced data such as a heatmap with a small Gaussian ellipsoid. It is defined as $\mathcal{FL}(p_t) = -\alpha_t (1 - p_t)^\gamma \log(p_t)$ with
\[
    p_t = \begin{cases}
p & \text{if } y = 1 \\
1 - p & \text{if } y = 0
\end{cases}
\]
    \item \textbf{Kullback-Leibler Divergence (KLDiv)}: Measures the relative entropy between two probability distributions, often used to compare predicted and target distributions in probabilistic models, and is defined as $D_{\text{KL}}(P \parallel Q) = \sum_{i} P(i) \log\left(\frac{P(i)}{Q(i)}\right)$
\end{itemize}

\noindent As noted earlier, a heatmap might be more expressive and carries more information, but if one or several specific position vectors are required for downstream tasks or evaluation metrics, a sampling approach needs to be chosen. Towards this end, two sampling strategies were explored:

\begin{itemize}
    \item \textbf{Argmax}: Picking the pixel with the highest probability is the simplest solution for single-endpoint prediction but is limited to a single prediction and is susceptible to noise. 
    \item \textbf{Clustering}: A clustering approach allows for controlled $n$-endpoint prediction, respecting larger blobs of prediction. 
    We picked DBSCAN~\cite{10.5555/3001460.3001507} with an epsilon of 3 and a minimum sample number of 10 and used the pixel with the maximum probability per cluster as its center point.
\end{itemize}

\noindent Since the training dataset and sampled situations contain many situations where the vehicle will not move, a typical prediction will contain two distribution clusters: one for the case that the vehicle is moving and one for the case that the vehicle remains at the same position. A typical prediction is shown in Figure~\ref{fig:prediction1}.

\subsection{Training}

The model is trained on the entire dataset of 2,190,000 battles, excluding 1,000 battles for evaluation during training and excluding another 100,000 battles for testing and final evaluation. It is trained for 20,000 steps for metric computations, and 100,000 steps for the final model used for visualization, with a batch size of 64.

\section{Evaluation}
    \label{sec:evaluation}

For the sake of comparability, we use the well-known final displacement error (FDE), defined as the Euclidean distance between the predicted and the true position, to measure the correctness of endpoint position predictions. To account for the probabilistic nature of location predictions, the minimum FDE of up to three samples is used (referred to as FDE@3).
However, as pointed out by multiple authors including Kumichev~\cite{Kumichev}, such simplifications do not always capture the true performance. A slightly better metric is the FDE relative to a baseline (\emph{Rel. FDE}), in our case the distance to the current vehicles position. This metric is able to provide more meaningful values for often occurring situations where, even though we explicitly sampled for moving vehicles, the average movement is very slow and vehicles tend to remain in close proximity to the original position. Defined as in Equation~\ref{eq:rel_fde}, a value of 1.0 would imply baseline behavior, with a value of 2.0 being twice as good.
\begin{equation}
    \text{Rel. FDE} = \frac{\text{FDE}(\text{current}, \text{target})}{\text{FDE}(\text{prediction}, \text{target})}
    \label{eq:rel_fde}
\end{equation}

\subsection{Loss Functions}

\begin{figure}[h]
    \centering
    \begin{minipage}{0.24\textwidth}
        \centering
        Focal Loss\\[3pt]
        \includegraphics[width=\linewidth]{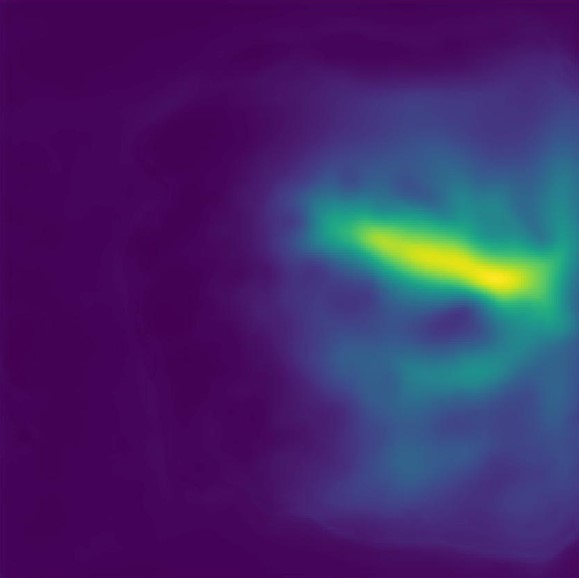}
    \end{minipage}
    \hfill
    \begin{minipage}{0.24\textwidth}
        \centering
        KLDiv\\[3pt]
        \includegraphics[width=\linewidth]{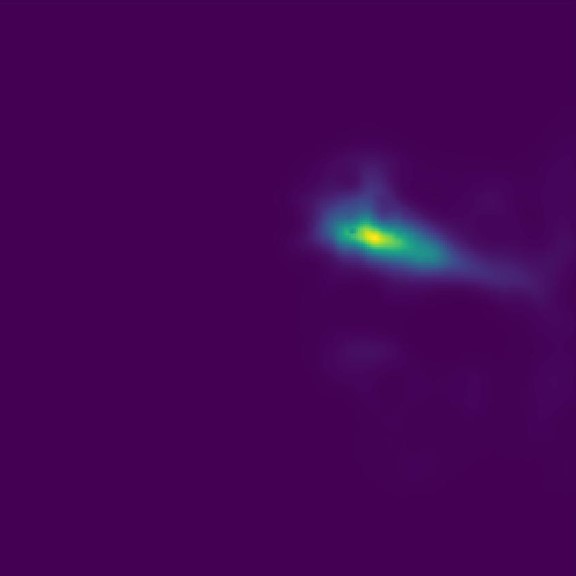}
    \end{minipage}
    \caption{Comparison between the two best performing loss functions.}
    \label{fig:losses}
\end{figure}

We evaluated the performance on all maps, with different environmental characteristics. Table~\ref{tab:loss} compiles the evaluation metrics for different loss functions, for both single sample FDE@1 and the clustered FDE@3. BCE and MSE perform worst, while focal loss performs similar to KLDiv for a single sample, but not for multi-sampling, where KLDiv outperforms all other loss functions by a large margin. When inspecting the output distribution as visualized in Figure~\ref{fig:losses}, focal loss tends to output very large distributions, and while the maxima are similar to those of KLDiv, it is overall less suitable as an approximation of a probability distribution or for clustering.

\begin{table}[t]
    \centering
    \caption{Evaluation results of different loss functions.}
    \label{tab:loss}
    \begin{tabular}{ccc}
         \toprule
         \textbf{Loss Function} & \textbf{Rel. FDE} & \textbf{Rel. FDE@3} \\
         \midrule
         BCE & 1.11 & 1.33 \\
         MSE & 1.18 & 1.30 \\
         Focal Loss & 1.32 & 1.35 \\
         KLDiv & 1.33 & 1.78 \\
         \bottomrule
    \end{tabular}
\end{table}

\subsection{Image Features}

Table~\ref{tab:features} contains the comparison between how the image feature is constructed in Section~\ref{sec:image_input}. Not including any additional features in the image leads to a lower performance. The encoded icons perform very similarly to the stacked Gaussian ellipsoids. As such and given their lower resource requirements due to the lower feature depth (and therefore lower trainable parameters) in the first image encoder layer, they are picked for all following experiments.

\begin{table}[ht]
    \centering
    \caption{Comparing the final performance between rendered vehicle icons and several stacked Gaussian representations.}
    \label{tab:features}
    \begin{tabular}{l c}
        \toprule
        \textbf{Approach} & \textbf{Rel. FDE@3} \\
        \toprule
        None & 1.52 \\
        Rendered Map & 1.78 \\
        Stacked Gaussian Ellipsoids & 1.74 \\
        \bottomrule
    \end{tabular}
\end{table}

\subsection{Architecture}

Table~\ref{tab:ablation} shows the average performance across all maps with different component combinations. We start with just the backbone \emph{U-Net} using the previously selected pre-trained \emph{EfficientNet-B2} architecture as the baseline. Then all components discussed above are successively added and evaluated, iteratively increasing the architecture’s complexity.

Looking at Table~\ref{tab:ablation}, one can see incremental improvement of the performance relative to the \emph{U-Net} baseline. Despite that, the total trainable parameters do not increase significantly because the \emph{U-Net} itself is the heaviest component. \new{It is also important to note that the image used as input for the \emph{U-Net++} already contains information about all vehicle positions, dampening the effect of the added components slightly.}

\subsection{Scenarios}

\begin{figure}[t]
    \centering
    \begin{minipage}[t]{0.233\textwidth}
        \vskip0pt
        \centering
        \includegraphics[width=\linewidth]{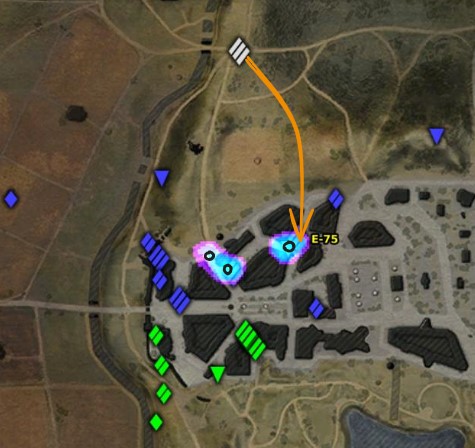}
        \caption{Three equally likely positions behind buildings (visualized as a superimposed heatmap) have been predicted (marked with black circles), with the prediction on the right being ultimately chosen based on the opponents teams advance.}
        \label{fig:prediction1}
    \end{minipage}
    \hfill
    \begin{minipage}[t]{0.24\textwidth}
        \vskip0pt
        \centering
        \includegraphics[width=\linewidth]{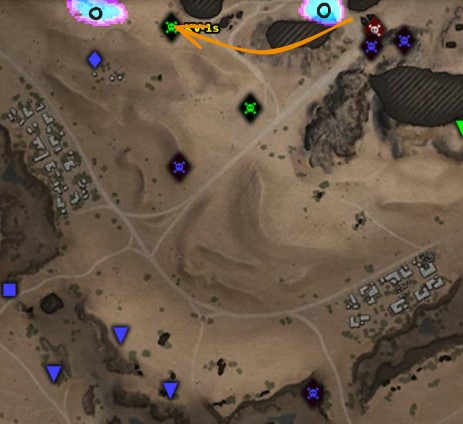}
        \caption{An notable prediction where the vehicle got destroyed while trying to reach the predicted position.}
        \label{fig:prediction2}
    \end{minipage}
\end{figure}

To further understand how the model predicts different scenarios, the best performing model (with all components applied) is inspected in more detail by visualizing example scenarios. One can observe smaller errors in urban maps due to the lower degree of freedom caused by the obstructed parts of the map. Figure~\ref{fig:prediction1} shows an example prediction leading into a more urban area, with three likely endpoints covered by buildings.

\begin{figure}[t]
    \centering
    \begin{minipage}[t]{0.21\textwidth}
        \vskip0pt
        \centering
        \includegraphics[width=\linewidth]{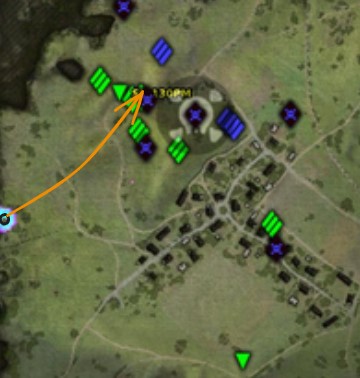}
        \caption{An example of a hard-to-predict behavior for an artillery which is leaving its safe spot and dashing into close combat against common reasoning.}
        \label{fig:prediction3}
    \end{minipage}
    \hskip5pt
    \begin{minipage}[t]{0.228\textwidth}
        \vskip0pt
        \centering
        \includegraphics[width=\linewidth]{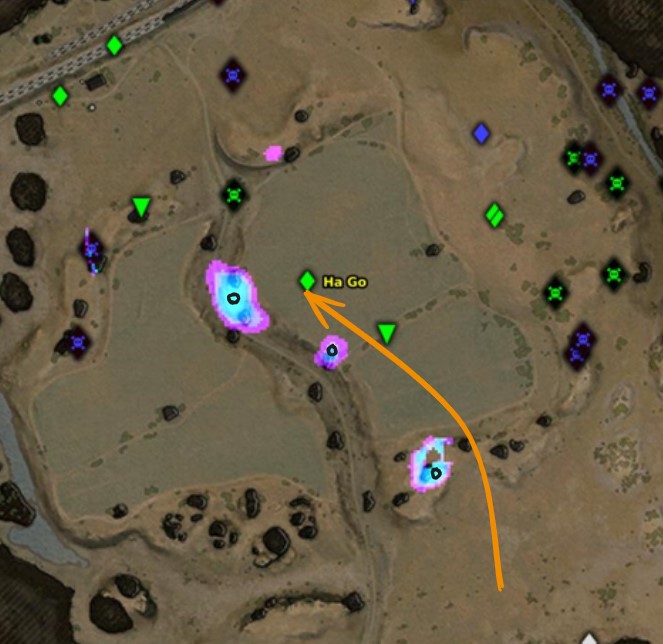}
        \caption{A prediction in an open area. The thresholded prediction shows several strategic points but the player instead took a path straight over the field.}
        \label{fig:prediction4}
    \end{minipage}
\end{figure}

An expected weakness in the model can be observed in scenarios where the player behaves unreasonably, for example, in Figure~\ref{fig:prediction3}, where the player is using a long range vehicle, yet drives towards the center and thus exposes themselves to enemy fire. Simultaneously, this is an example of how anomaly detection could be implemented. Another expected issue is large open spaces, as seen in the example visualized in Figure~\ref{fig:prediction4}, where the model has to spread the probability distribution over a wide area, preventing it from showing up as peaks. A third weakness can occur with rapid and rare changes in conditions such as illustrated in Figure~\ref{fig:prediction2} where the target vehicle got shot and destroyed in the middle of transit.

\begin{table}[t]
    \centering
    \caption{Ablation study on the different components presented in Section~\ref{fig:architecture}, with their respective size in millions (M) of parameters and evaluated Rel. FDE.}
    \label{tab:ablation}
    \begin{tabular}{lcc}
        \toprule
        \textbf{Components} & \textbf{M Params} & \textbf{Rel. FDE@3} \\
        \midrule
        \emph{U-Net++} without any context & 28.914 M & 1.66 \\
        With replay context & 28.915 M & 1.70 \\
        With the target vehicle as context & 28.975 M & 1.74 \\
        With all vehicles as context & 29.160 M & 1.78 \\
        \bottomrule
    \end{tabular}
\end{table}

\subsection{Vehicle Types}

Another interesting dataset split is by tank type, since different types require different strategies. Table~\ref{tab:vehicle_types} showcases the distribution of speed, where especially high variance is harder to predict, as well as their Rel. FDE average across all maps. Especially artillery is harder to predict better than the baseline since they remain on the same spot for a long time, but react to shifts on a higher strategic level.

\begin{table}[t]
    \centering
    \caption{WoT vehicle types with speed characteristics and Rel. FDE.}
    \label{tab:vehicle_types}
    \begin{tabular}{l cc c}
        \toprule
        \textbf{Vehicle Type} & \multicolumn{2}{c}{\textbf{Speed (m/s)}} & \textbf{Rel. FDE@3} \\
        & \textbf{Avg.} & \textbf{Std.} & \\
        \midrule
        Light Tank      & 2.385 & 10.971 & 1.461 \\
        Medium Tank     & 2.236 & 8.304  & 1.663 \\
        Heavy Tank      & 2.606 & 6.454  & 1.764 \\
        Tank Destroyer  & 1.524 & 5.106  & 1.570 \\
        Artillery       & 0.545 & 1.667  & 1.036 \\
        \bottomrule
    \end{tabular}
\end{table}

\section{Discussion and Future Work}
    \label{sec:discussion}

In this work, we have presented a multimodal architecture utilizing various techniques to improve location prediction performance on top of \emph{U-Net} and provided a foundation for further research. It has also shown some weaknesses, such as underwhelming representation of certain predictions with endpoints spread over larger areas and with respect to illogical outliers, which should be addressed in the future.

Further research can be conducted on how to optimize the image-data, that is the input to the \emph{U-Net}, to best represent the game scenario. As shown with the Gaussian distribution image inputs stacked on top of the RGB map, it is possible to include and stack further static image data (e.g., collision information, height maps, concealment information) to make use of the presented dimensional scalability. Also, there are many additional image encoding architectures which could be tested such as \emph{Segformers}~\cite{xie2021segformersimpleefficientdesign} which uses hierarchical attention to, among other benefits, achieve higher global awareness.

Additionally, while our current research focuses solely on the task of endpoint location prediction, the multimodal architecture presented can be adapted and used for various other tasks, possibly even being suitable to multi-task or multi-agent outputs. Currently, the final embedding vector is used to condition the primary image prediction decoder, but it also provides a solid starting point for other types of predictions, clustering, regression, and classification tasks. In this context, the embedding vector could be used to encode the entire game situation in a compact data representation. Going further in this direction, several such prediction heads can be built upon this final embedding, harnessing the benefits of multi-task learning such as improved generalization capability and more efficient use of training data by reducing overfitting. Similar to multi-task outputs, predictions for multiple agents could also be done at once without further adjustments. In this paper, a final cross-attention mechanism with the target vehicle is used to project to a single embedding vector. However, for some tasks it could be beneficial to have one vector per input vehicle. This allows for multi-agent predictions without further adjustments.

To retain more control over the prediction outputs, prediction can be conditioned on additional inputs such as done by Wang et al.~\cite{tacticai}, retrieving predictions given a certain expected outcome. We implicitly make use of a simplified version of this technique by conditioning the output on the expected prediction time horizon. However, this can be extended to, for example, adjusting the behavior of one team to maximize the probability of winning.

In terms of the data used as the basis for predictions, currently for simplicity we ignore visibility and predict on the fully observable game state, as opposed to using the last known position from the viewpoint of, for example, the target vehicle which position is being predicted. However, especially for real-time prediction tasks, a fair game state representation with some enemy vehicles being hidden and represented by their last known state, would be more appropriate. Another improvement for the underlying battle data could be the usage of a significantly higher step count during training to further improve accuracy, especially given the relatively large dataset. Even at a comparatively small step count, the validation shows consistent results to accurately compare different configurations. Lastly, as an improvement for the target position representation, in environments with fine granular obstacles (e.g., buildings) the Gaussian ellipsoid may bleed to the opposite side, despite being physically impossible to reach. Therefore, instead of simply masking out obstacles, the blob should be generated in a flood fill or simulative manner to respect physical constraints.

\section{Conclusions}
    \label{sec:conclusions}

In this paper, we proposed a team-based endpoint prediction architecture using a multimodal encoder and an enhanced \emph{U-Net} model architecture as a backbone. By integrating spatial and temporal data, conditioning outputs on prediction horizons, and employing attention mechanisms, location accuracy can be noticeably improved. Our evaluation demonstrates the impact of the various architectural components, highlighting the effectiveness of incorporating vehicle context, history, and global game state information. Our findings establish a strong foundation for future research, including multi-agent prediction, additional feature integration, and real-time inference with partial observability.

\section*{Acknowledgments}

We would like to thank \emph{Wargaming} for providing a high-quality dataset to experiment and evaluate on.

\bibliographystyle{IEEEtran}
\bibliography{references}

\end{document}